\documentclass[conference]{IEEEtran}
\usepackage{amsmath}
\usepackage{amssymb}
\usepackage{booktabs}
\usepackage{graphicx}
\newcommand{\pmm}[1]{\ensuremath{\,{\scriptstyle\pm #1}}}

\ifCLASSINFOpdf
\else
\fi
\begin{document}
%
\title{Uncertainty-Aware Learning from Multi-Expert Interval Targets }

\makeatletter
\newcommand{\linebreakand}{%
  \end{@IEEEauthorhalign}\hfill\mbox{}\par
  \mbox{}\hfill\begin{@IEEEauthorhalign}
}
\makeatother

\newcommand{\authblock}[1]{%
  \IEEEauthorblockA{\parbox{9cm}{\centering #1}}%
}

\author{
\IEEEauthorblockN{Samira Alkaee Taleghan}
\authblock{
\textit{University of Colorado Denver}\\
Denver, Colorado, USA\\
samira.alkaeetaleghan@ucdenver.edu}
\and
\IEEEauthorblockN{Younghyun Koo}
\authblock{
\textit{National Snow and Ice Data Center (NSIDC)}\\
\textit{CIRES, University of Colorado Boulder}\\
Boulder, Colorado, USA\\
Younghyun.Koo@colorado.edu}

\linebreakand

\IEEEauthorblockN{Andrew P. Barrett}
\authblock{
\textit{National Snow and Ice Data Center (NSIDC)}\\
\textit{CIRES, University of Colorado Boulder}\\
Boulder, Colorado, USA\\
andrew.barrett@colorado.edu}
\and
\IEEEauthorblockN{Farnoush Banaei-Kashani}
\authblock{
\textit{University of Colorado Denver}\\
Denver, Colorado, USA\\
farnoush.banaei-kashani@ucdenver.edu}
}


%


\maketitle

\begin{abstract}
Many machine learning (ML) applications rely on expert labels, and qualified experts may provide different but plausible interpretations of the same observation. Such variation across expert labels may reflect genuine disagreement or ambiguity rather than annotation error. When individual experts additionally report intervals rather than exact values, the supervision contains two distinct sources of label uncertainty: within-label imprecision and between-expert variation.
Existing methods treat these forms separately: multi-expert approaches
collapse labels to a consensus, interval-target methods often yield a single prediction, and
predictive-uncertainty methods rarely validate their uncertainty estimates
against observed expert disagreement. 
To address this problem, we propose an approach that preserves individual expert intervals, separates within-label imprecision from between-expert variation, and validates the corresponding predictive uncertainty components. First, heterogeneous
label vocabularies are harmonized into a common probabilistic label space,
separating encoding differences from expert judgement. Second, individual
label intervals are retained and modeled with a mixture of Beta distributions
trained using a proper Cram\'er-distance objective, preserving distinct expert-reported labels. Third, we decompose predictive uncertainty into within-component, between-component, and model uncertainty, and evaluate whether these components correspond to within-label uncertainty, between-label uncertainty, and model error, respectively. Because this correspondence is not guaranteed, we introduce decomposition matching, which aligns the predictive components to their intended label-side sources. On sea-ice concentration (404 scenes labeled by multiple ice analysts), the model reduces MAE by 31\% over hard labels and outperforms aggregation, interval-distribution, and interval-regression baselines. Uncertainty components do not automatically align with their sources: decomposition matching raises the correlation with between-expert variation to 0.500 while reducing correlation with model error, with essentially unchanged point accuracy.
\end{abstract}

%
\IEEEpeerreviewmaketitle


\section{Introduction}
\label{sec:intro}
Many machine-learning applications rely on expert labels, where the target itself is a matter of interpretation rather than a directly observed quantity. Consequently, qualified experts may provide different but plausible labels for the same instance, introducing uncertainty in the supervision itself rather than only in the predictive model. Radiologists may delineate the same lesion differently~\cite{kohl2018probabilistic,warfield2004staple}, annotators may provide different judgments on the same item~\cite{dawid1979maximum,raykar2010learning}, and institutional analysis pipelines may produce different labels for the same observation. Such differences do not necessarily indicate annotation error. Plank~\cite{plank2022problem} described human label variation as arising from genuine disagreement, subjectivity, annotator uncertainty, or multiple plausible interpretations. Preserving this variation can therefore be important: it identifies cases where the available evidence supports more than one interpretation, where predictions may warrant additional review, and where another expert assessment may be particularly informative.

The problem becomes richer when individual labels are themselves
imprecise. An expert may report an interval of plausible values rather than a
precise point, reflecting limited precision or ambiguity in the judgement.
With multiple experts, this creates two distinct forms of uncertainty in the
labels: uncertainty within an individual label, represented by the range
of values that expert considers plausible, and uncertainty between
labels, arising when different experts report different intervals for the same
instance. A learned predictor introduces a third source of uncertainty: uncertainty in the model itself. 
These sources reflect different phenomena and warrant separate treatment. A learning framework for this setting should distinguish
within-label uncertainty, between-label uncertainty, and model
(epistemic) uncertainty rather than collapsing them into a single target or a
single uncertainty measure.
Existing approaches address parts of this problem separately. Multi-expert methods typically infer a consensus or latent ground truth~\cite{warfield2004staple,dawid1979maximum}, model expert reliability~\cite{raykar2010learning,rodrigues2018deep}, or simply select one trusted label~\cite{stokholm2024autoice}. Other work preserves label variability through label distributions~\cite{geng2016label,peterson2019human} or probabilistic structured prediction~\cite{kohl2018probabilistic}, while predictive-uncertainty methods quantify model uncertainty using ensembles, Bayesian approximations, or evidential prediction~\cite{lakshminarayanan2017simple,gal2016dropout,sensoy2018evidential,amini2020deep}. Recent interval-supervision methods preserve bounded targets during training, but largely focus on deterministic regression~\cite{cheng2023weakly,pukdee2025learning}.
However, these approaches do not jointly distinguish and validate label imprecision, between-expert variation, and model uncertainty using observed multi-expert interval labels.

In this work, we propose a framework for learning from heterogeneous, interval-valued expert labels. We construct parallel label-side and predictive variance decompositions and examine whether the predictive components correspond to the observed sources of label uncertainty. First, we harmonize the labels into distributions over a common label space, separating encoding differences from differences in judgement. Second, we retain the individual label intervals and learn a mixture of Beta distributions~\cite{bishop1994mixture} using a proper Cram\'er-distance objective~\cite{bellemare2017cramer}, allowing distinct expert interpretations to be represented. Third, we evaluate whether uncertainty within and between mixture components, together with model uncertainty estimated using a deep ensemble of independently trained models, correspond to label imprecision, between-expert variation, and model error, respectively, and introduce decomposition matching to explicitly align the predictive components with their intended label-side sources.

We study this problem through sea-ice concentration (SIC) estimation, an operational setting in which supervision is both interval and provided by multiple experts. SIC describes the fraction of an area covered by sea ice and is important for climate monitoring, numerical modeling, maritime navigation, and decision-making in polar regions \cite{Vihma,taleghan2024semisupervised}. Operational SIC information is commonly provided through ice charts produced by national ice services, such as the U.S. National Ice Center (NIC), whose trained analysts interpret satellite observations and manually delineate regions of relatively homogeneous ice conditions. This interpretation is not always unique: SIC can be difficult to assess under intermediate concentrations, marginal ice conditions, poorly resolved floes, and melt-season imagery \cite{cheng2020accuracy}, so qualified analysts may produce different but plausible interpretations of the same observation. Each chart polygon is described using the standardized World Meteorological Organization (WMO) Egg Code and SIGRID3 data format \cite{NSIDC,jalayer2025}, and these chart-derived labels are widely used to train SIC prediction models \cite{alkaeetaleghan2025icebench,Taleghan2025icefmbench}. In this work, we treat each ice service as a distinct source of expert labels and use independently produced charts to characterize expert label variation. The resulting supervision exposes three distinct sources of uncertainty: an individual SIC label may span a range of plausible values, producing within-label imprecision; different ice services may report different intervals for the same observation, producing between-expert variation; and the learned predictor introduces model uncertainty. We study these sources using 404 satellite scenes independently charted by up to three operational ice services.

We make three main contributions:
(i) we formulate learning from multiple interval labels while
explicitly separating within-label uncertainty from between-expert variation, and harmonize heterogeneous label vocabularies into a common
probabilistic label space;
(ii) we construct a multi-expert interval target and train a
Mixture-of-Betas predictor using a proper Cram\'er objective; and
(iii) we decompose predictive uncertainty into within-component,
between-component, and model uncertainty, validate each against
within-label uncertainty, between-expert variation, and model error,
respectively, and show that between-component uncertainty reliably
recovers between-expert variation only under explicit decomposition matching.

The remainder of the paper is organized as follows. Section II reviews related work, Section III presents the methodology, Section IV describes the experimental setup and reports the results, including sensitivity to key design choices, and Section V concludes.

\section{Related Work}
\label{sec:related}

\subsection{Learning from Multiple Experts}

\paragraph{Consensus and Expert Reliability}

Learning from multiple experts traditionally formulated as a
problem of recovering a latent ground-truth label from noisy or inconsistent
observations. Classical approaches estimate expert reliability together
with a consensus label
\cite{dawid1979maximum,warfield2004staple,raykar2010learning}, while later
methods incorporate expert-specific confusion or bias directly into neural
models \cite{rodrigues2018deep}. These approaches are effective when
disagreement primarily reflects differences in expert reliability, but
their objective is generally to resolve disagreement in order to recover a
single underlying target.

\paragraph{Learning from Expert Disagreement}

A complementary line of work treats disagreement as informative rather than as annotation noise. Methods preserve multiple annotations or model individual annotators to capture variation in expert judgement \cite{uma2021learning,davani2022dealing}, while Plank~\cite{plank2022problem} argued that human label variation may reflect genuine ambiguity or differences in interpretation. Other approaches represent this variation through label distributions \cite{geng2016label,peterson2019human} or multiple plausible structured outputs \cite{kohl2018probabilistic}.
These methods generally consider point-valued annotations, with label distributions used to represent variation across individual labels. Our setting instead considers multiple interval expert labels, allowing us to represent both imprecision within each expert label and variation across expert labels. We preserve these sources in the target distribution, evaluate whether the corresponding predictive uncertainty components recover them, and use decomposition matching when the between-component uncertainty does not emerge reliably.

\subsection{Learning from Interval Targets}

Interval supervision is approached from several perspectives.
Earlier work on interval data prediction learns interval
outputs from interval observations, including center--range regression,
constrained interval regression, kernel methods, and neural approaches
\cite{neto2008centre,neto2010constrained,
sadeghi2019efficient}. Related statistical work on
interval-censored regression similarly learns from outcomes known only to lie
within bounds \cite{rabinowitz1995regression}.
More recently, regression with interval targets (RIT) considers the setting in
which an interval contains an unknown point-valued target.~\cite{cheng2023weakly} formulate RIT as weakly supervised regression and
constrain predictions using the observed interval.~\cite{pukdee2025learning} further develop interval-compatible and
worst-case objectives for this setting.
These methods generally consider a single interval target. Our setting instead retains multiple expert-reported intervals for the same observation, allowing within-label imprecision to be distinguished from between-expert variation and represented separately in the predictive uncertainty.

\subsection{Probabilistic Prediction and Uncertainty}

Predictive-uncertainty methods quantify uncertainty arising from the data,
the model, or both. Approximate Bayesian inference through Monte Carlo dropout
\cite{gal2016dropout}, deep ensembles
\cite{lakshminarayanan2017simple}, and evidential approaches
\cite{sensoy2018evidential,amini2020deep} provide widely used mechanisms for
estimating predictive or epistemic uncertainty. These methods identify uncertain predictions, but the uncertainty reflects model behavior rather than observed between-expert variation. High predictive variance alone cannot distinguish label imprecision, between-expert variation, or missing model information.
Representing multiple expert interval labels requires modeling both bounded uncertainty and multiple plausible outcomes. Beta distributions are well suited to bounded targets on $[0,1]$ \cite{ferrari2004beta}, but a single Beta cannot represent separated modes. Mixture models \cite{bishop1994mixture} address this limitation. Our Mixture-of-Betas combines these ideas and tests whether its uncertainty components correspond to within-label imprecision, between-expert variation, and model uncertainty.

\subsection{Uncertainty in SIC Estimation}

Operational ice charts provide expert-derived estimates of SIC from satellite
observations. Because SIC is assigned through analyst interpretation, estimates
for the same conditions can vary across analysts and operational ice services.  Previous studies quantified variability in these assessments
both among individual ice analysts
\cite{karvonen2015comparison,cheng2020accuracy}
and across operational ice services
\cite{panchi2022assessment}. These studies show that differences in SIC estimates vary with
concentration, location, and season, with larger differences at intermediate
SIC values.
For SIC estimation, uncertainty mainly quantified from the predictive
model. Heteroscedastic Bayesian neural networks used to separate
aleatoric uncertainty, represented as input-dependent predictive variance, from
epistemic uncertainty arising from uncertainty in the model parameters
\cite{chen2023predicting}. Other work derives SIC and its predictive uncertainty
from calibrated ensemble probabilities over discrete concentration levels
\cite{wulf2024panarctic}. In these approaches, aleatoric or predictive
uncertainty is inferred from the model output rather than measured from
independent expert labels.
In contrast, our work explicitly models uncertainty in the SIC supervision
itself. We preserve interval labels from multiple experts, distinguish within-label imprecision from between-expert
 variation, and validate whether the learned predictive uncertainty
corresponds to these observed sources of label uncertainty.

\section{Methodology}
\label{sec:method}
The methodology consists of three stages: harmonizing heterogeneous label codes into a common label space, constructing a probabilistic target that preserves within-label imprecision and between-expert variation, and decomposing predictive uncertainty into components that can be compared with these observed label uncertainties and model error.
\subsection{Problem Formulation and Label Harmonization}

Let $x\in\mathcal{X}$ denote an instance, here a pixel together with its input
features, and let the ordered label space be $\mathcal{Y}=[0,1]$. Each instance is labeled by a subset $\mathcal{A}(x)\subseteq\{1,\dots,S\}$ of a pool of $S$ experts, with
$A(x)=|\mathcal{A}(x)|\geq1$ varying across instances. Expert $a$ reports a
label encoded as code $c_a$ from a vocabulary $\mathcal{V}_a$, which may differ
across experts. A decoding map
\begin{equation}
\iota_a:\mathcal{V}_a
\rightarrow
\{[L,U]:0\leq L\leq U\leq1\}
\end{equation}
maps each code to an interval
$[L_a,U_a]=\iota_a(c_a)$, where $L_a=U_a$ denotes an exact-valued label.
The observed label set for $x$ is therefore
\begin{equation}
\mathcal{D}(x)
=
\bigl([L_a,U_a]\bigr)_{a\in\mathcal{A}(x)},
\label{eq:label_set}
\end{equation}
and the training set is
$\{(x_i,\mathcal{D}(x_i))\}_{i=1}^{N}$.
This label set contains two forms of label uncertainty: within-label
uncertainty, represented by the range of plausible values within each label
reported by an expert, and between-label uncertainty, induced by differences
among the labels reported by different experts for the same instance.

Because experts use different coding vocabularies and quantization
conventions, the raw codes are not directly comparable. We therefore treat
each label as uniform on its interval and discretize it onto a common grid:
partition $[0,1]$ into $J$ bins $\mathcal{B}_j=[\ell_j,u_j)$ of width $1/J$
with centers $z_j$, the final bin closed. For $L_a<U_a$, the unnormalized
mass in $\mathcal{B}_j$ is the overlap length
\begin{equation}
\tilde p_a(j)
=
\max\bigl(0,\;\min(U_a,u_j)-\max(L_a,\ell_j)\bigr),
\end{equation}
normalized by $\sum_{j'}\tilde p_a(j')=U_a-L_a$; for an exact label
$L_a=U_a$ all mass goes to the bin containing it. This yields a harmonized
distribution $p_a$ for each reported label.
The harmonized distributions $p_a$ provide a common representation for
comparing labels expressed under heterogeneous coding conventions, whereas
the original intervals $[L_a,U_a]$ are retained for distributional learning
and for quantifying within-label uncertainty and between-label uncertainty. In particular, between-label uncertainty arises from variation among the
labels reported by different experts for the same instance. 
To quantify the
magnitude of this between-expert variation, we use the mean pairwise Wasserstein-1 distance. For brevity, let $A=A(x)$:
\begin{equation}
d_{\mathrm{src}}(x)
=
\frac{2}{A(A-1)}
\sum_{a<b}
W_1(p_a,p_b),
\qquad A\geq2,
\label{eq:w1_disagreement}
\end{equation}
where $W_1$ is computed on the ordered label grid using the bin locations as
the ground metric. When $A(x)=1$, between-expert variation cannot be assessed, and
$d_{\mathrm{src}}$ is therefore undefined rather than zero.
We additionally define two
forms of uncertainty directly from the reported intervals. Let \(m_a=(L_a+U_a)/2\), and \(\bar m=\frac{1}{A}\sum_{a\in\mathcal{A}(x)}m_a\).
Treating each label as uniform on its interval, the equal-weight mixture
$\tfrac{1}{A}\sum_{a\in\mathcal{A}(x)}\mathrm{Unif}[L_a,U_a]$ has variance
$t_w^2(x)+t_b^2(x)$ by the law of total variance, where
\begin{equation}
\begin{aligned}
t_w^2(x)
&=
\frac{1}{A}
\sum_{a\in\mathcal{A}(x)}
\frac{(U_a-L_a)^2}{12},
\qquad A\geq1,\\
t_b^2(x)
&=
\frac{1}{A}
\sum_{a\in\mathcal{A}(x)}
(m_a-\bar m)^2,
\qquad A\geq2.
\end{aligned}
\label{eq:label_var}
\end{equation}
Here, $t_w^2$ measures within-label uncertainty, defined as the average
within-label variance induced by the widths of the reported intervals, while
$t_b^2$ measures between-label uncertainty through the variance of their
interval centers. The $1/A$ normalization makes
$t_b^2$ the between-label component of the variance of the equal-weight
label mixture. These quantities characterize uncertainty observed from the expert labels.
Model uncertainty, which arises from variability in the learned predictor, is
estimated separately using a deep ensemble.

\subsection{Distributional Learning and Uncertainty Decomposition}
\label{sec:distributional}
The quantities in \eqref{eq:label_var} describe within-label and between-label
uncertainty, but these distinctions must also be preserved in the training
target. We therefore construct the target from the reported intervals
by representing each as a probability distribution.
For expert $a$, the reported interval $[L_a,U_a]$ is treated as a uniform
distribution, assigning equal probability density to all values within the
reported range. Let $Q_a(y)$ denote its cumulative distribution function
(CDF). Then
\begin{equation}
Q_a(y)=
\begin{cases}
\mathrm{clip}\!\left(
\dfrac{y-L_a}{U_a-L_a},0,1
\right),
& U_a>L_a,\\[4pt]
\mathbb{I}[y\geq L_a],
& U_a=L_a.
\end{cases}
\label{eq:annotation_cdf}
\end{equation}
When $U_a>L_a$, the CDF increases from $0$ to $1$ across the reported
interval; when $U_a=L_a$, the label is exact and the distribution reduces to
a point mass at $L_a$.
For an instance with $A$ available expert-reported labels, the target CDF gives
each label equal weight:
$Q(y\mid x)=A^{-1}\sum_{a\in\mathcal{A}(x)}Q_a(y)$.
For interval labels, the corresponding density is
$q(y\mid x)=A^{-1}\sum_{a\in\mathcal{A}(x)}
\mathrm{Uniform}(y;L_a,U_a)$.
This construction retains the width of each reported interval and the
differences among the intervals reported by different experts. Overlapping intervals
reinforce common support, while separated intervals remain separated rather
than being merged into a single interval.

The constructed target may contain both broad intervals and separated regions
of probability when different experts report conflicting intervals. The predictive model therefore
needs a distribution that can represent this structure while respecting the
bounded label space $\mathcal{Y}=[0,1]$. A Beta distribution is a continuous
distribution on $[0,1]$, parameterized by two positive shape parameters
$\alpha$ and $\beta$, making it suitable for modeling bounded label space.
A single Beta distribution, however, cannot represent multiple separated
high-probability regions without also assigning probability to the values
between them. We therefore model the predictive distribution as a mixture of
Beta distributions \cite{bishop1994mixture},
\begin{equation}
p_\theta(y\mid x)
=
\sum_{k=1}^{K}
\pi_k(x)\,
\mathrm{Beta}
\bigl(y;\alpha_k(x),\beta_k(x)\bigr),
\label{eq:beta_mixture}
\end{equation}
where $\pi_k(x)\geq0$ and $\sum_k\pi_k(x)=1$, and
$\alpha_k(x),\beta_k(x)>0$ determine the shape of each component. The mixture
allows different components to represent different regions favored by the
reported labels while maintaining a continuous predictive distribution over $[0,1]$.

We train the predictive distribution to match the full target constructed from
the label set $\mathcal{D}(x)$ rather than a point summary. Let $F_\theta(y\mid x)$ denote the predictive CDF, and we minimize the Cram\'er distance between $F_\theta$ and the target CDF:
\begin{equation}
D_{\mathrm{Cr}}(F_\theta,Q)
=
\int_0^1
\left[
F_\theta(y\mid x)-Q(y\mid x)
\right]^2dy .
\label{eq:cramer}
\end{equation}
This objective is closely related to the Continuous Ranked Probability Score
(CRPS), a proper scoring rule for probabilistic predictions. Specifically, the
expected CRPS for $Y\sim Q$ differs from
$D_{\mathrm{Cr}}(F_\theta,Q)$ only by a term that depends on $Q$.
Therefore, minimizing \eqref{eq:cramer} is equivalent to minimizing expected
CRPS with respect to the target distribution \cite{gneiting2007strictly}.
This encourages the model to reproduce the full target distribution rather
than only its mean.

Beyond matching the target distribution, we also examine how uncertainty is
represented within the predictive mixture. Because the target uncertainty was
separated into within-label and between-label terms in
\eqref{eq:label_var}, we derive an analogous decomposition for the predictive
distribution. For mixture component $k$, let
\[
\mu_k=
\frac{\alpha_k}{\alpha_k+\beta_k},
\qquad
\sigma_k^2=
\frac{\alpha_k\beta_k}
{(\alpha_k+\beta_k)^2(\alpha_k+\beta_k+1)},
\]
and let $\mu=\sum_k\pi_k\mu_k$. By the law of total variance,
\begin{equation}
\mathrm{Var}(Y\mid x)
=
\underbrace{
\sum_k\pi_k\sigma_k^2
}_{s_w^2(x)}
+
\underbrace{
\sum_k\pi_k(\mu_k-\mu)^2
}_{s_b^2(x)} .
\label{eq:ltv}
\end{equation}
The first term, $s_w^2$, measures variation within predictive components,
whereas $s_b^2$ measures variation between component means. This decomposition
has the same within/between structure as the quantities derived from the
reported labels in \eqref{eq:label_var}: $t_w^2$ and $s_w^2$ describe
within-label and within-component uncertainty, respectively, while $t_b^2$ and
$s_b^2$ describe between-label and between-component uncertainty. We use a variance-based rather than entropy-based decomposition so that
$s_w^2$ and $s_b^2$ are directly comparable to the variances $t_w^2$ and $t_b^2$.
However, the Cram\'er loss constrains only the overall predictive distribution,
so this correspondence is not guaranteed.

Because this correspondence is not guaranteed by the Cram\'er objective, we
first evaluate whether it emerges in the unconstrained mixture. We then introduce decomposition matching, which adds auxiliary loss
terms that match the within- and between-component uncertainties to their
label-derived counterparts, $t_w^2$ and $t_b^2$.
Let $\mathcal{I}$ denote the set of valid training pixels and
$\mathcal{I}_2=\{x\in\mathcal{I}:A(x)\geq2\}$ denote those with at least two
expert-reported labels. The resulting objective is
\begin{equation}
\begin{split}
\mathcal{L}_{\mathrm{decomp}}
={}&
\left\langle
D_{\mathrm{Cr}}
\right\rangle_{\mathcal I}
+
\lambda_w
\left\langle
\left|
\sqrt{s_w^2+\epsilon}
-
\sqrt{t_w^2+\epsilon}
\right|
\right\rangle_{\mathcal I}
\\
&+
\lambda_b
\left\langle
\left|
\sqrt{s_b^2+\epsilon}
-
\sqrt{t_b^2+\epsilon}
\right|
\right\rangle_{\mathcal I_2},
\end{split}
\label{eq:decomp}
\end{equation}
where $\epsilon=10^{-6}$. The within-component term is aligned with
within-label uncertainty wherever a valid interval is available, whereas the
between-component term is aligned with between-label uncertainty only when at
least two labels are available. With the Cram\'er objective alone,
any correspondence arises naturally; under \eqref{eq:decomp}, it is explicitly
enforced during training.

Finally, we distinguish uncertainty represented within the predictive
distribution from uncertainty associated with the learned model. We train
$R$ independently initialized and trained models as a deep ensemble
\cite{lakshminarayanan2017simple}. For ensemble member $r$, let
$\pi_{rk}$, $\mu_{rk}$, and $\sigma_{rk}^2$ denote the weight, mean, and
variance of mixture component $k$, respectively. The predictive mean of
ensemble member $r$ and the ensemble mean are
$\mu_r=\sum_k \pi_{rk}\mu_{rk}$ and
$\bar{\mu}=\frac{1}{R}\sum_{r=1}^{R}\mu_r$.
Applying the law of total variance over both mixture components and ensemble
members gives
\begin{equation}
\begin{aligned}
\mathrm{Var}(Y\mid x)
={}&
\underbrace{
\mathbb{E}_r
\left[
\sum_k \pi_{rk}\sigma_{rk}^2
\right]
}_{u_w^2}
\\
&+
\underbrace{
\mathbb{E}_r
\left[
\sum_k \pi_{rk}
(\mu_{rk}-\mu_r)^2
\right]
}_{u_b^2}
\\
&+
\underbrace{
\frac{1}{R}
\sum_{r=1}^{R}
(\mu_r-\bar{\mu})^2
}_{u_m^2}.
\end{aligned}
\label{eq:three_way_decomp}
\end{equation}
where $\mathbb{E}_r[\cdot]$ denotes the average over ensemble members.
The three terms describe within-component, between-component, and model
(epistemic) variability, respectively. The model term $u_m^2$ measures
variation in the predictive mean across independently trained ensemble
members. On held-out data, we evaluate their
interpretation: $u_w^2$ is compared against the interval-derived within-label
uncertainty $t_w^2$, $u_b^2$ is compared against the between-label uncertainty
$t_b^2$ and the complementary measure of between-expert variation
$d_{\mathrm{src}}$, and $u_m^2$ is compared against prediction error.

\section{Experimental Evaluation}
\label{sec:experiments}

This section describes the multi-expert SIC dataset, experimental setup, comparison baselines, and evaluation results.
\begin{figure}[t]
\centering
\includegraphics[width=\columnwidth]{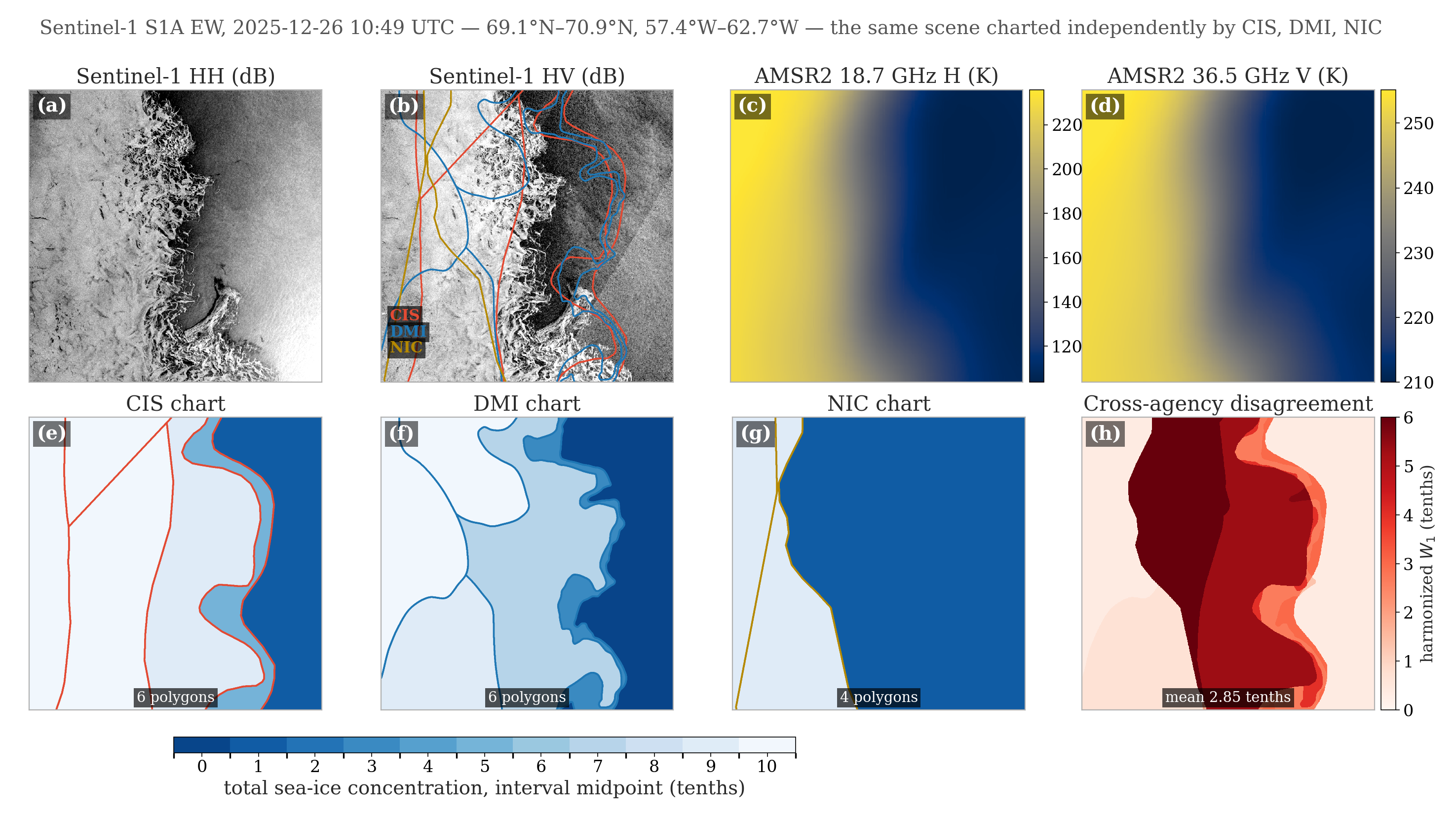}
\caption{One Sentinel-1 EW scene (2025-12-26, Davis Strait).
(a) HH backscatter and (b) HV backscatter with polygon boundaries from
CIS, DMI, and NIC. (c)--(d) AMSR2 18.7-GHz H- and 36.5-GHz V-polarized
brightness temperatures. (e)--(g) SIC charts from CIS, DMI, and NIC,
shown at the midpoint of each reported interval. (h) Mean pairwise $W_1$
distance among the available agency labels, concentrated near the
marginal ice zone.}
\label{fig:scene}
\end{figure}

\subsection{Dataset}
\label{sec:datasets}
We construct a multi-expert sea-ice concentration (SIC) dataset by matching the same satellite
observation with independently produced operational ice charts. Trained expert ice analysts construct each ice chart by delineating polygons
that represent regions of approximately homogeneous ice conditions and
assigning a sea-ice concentration (SIC) label to each polygon. SIC denotes the
fraction of the surface covered by sea ice, from 0 to 100\%, and is reported
using standardized World Meteorological Organization (WMO) codes that encode
concentration as an interval rather than a single exact value
~\cite{wmo2014sigrid3}. The labels are provided by four operational sources: the
U.S. National Ice Center (NIC), the Danish Meteorological Institute (DMI), the
Canadian Ice Service (CIS), and the National Oceanic and Atmospheric
Administration (NOAA). We treat each independently produced agency chart as
one expert-reported label set for the corresponding satellite observation.
The width of each SIC interval represents within-label uncertainty, while
differences among the intervals reported by different agencies for the same
observation represent between-label uncertainty. Figure~\ref{fig:scene} illustrates this setting for one
Sentinel-1 observation and the independently produced expert charts.

The agencies use different SIC vocabularies: NIC and NOAA predominantly use SIGRID-3 range codes, whereas DMI and CIS use decile-based intervals. We therefore harmonize the labels to a common 11-bin concentration representation before comparing expert reports (e.g., NIC 46, i.e.\ 40--60\%, vs.\ DMI 40 becomes a $W_1$ of 1.5
tenths after harmonization rather than a categorical mismatch). Table~\ref{tab:pairs} presents the resulting pairwise comparison across 404 distinct scenes, of which 364 are charted by two agencies and 40 by three agencies; one DMI–NIC scene-pair with no shared valid pixels is excluded. Both the frequency and magnitude of between-expert variation differ substantially across agency pairs. DMI--NIC, CIS--NIC, and NIC--NOAA show relatively large mean $W_1$ distances of 0.80--0.89 tenths, whereas CIS--NOAA shows substantially closer agreement. Across the reported label intervals, approximately 83\% of target variance is between experts and 17\% is within individual labels, providing label-side references for evaluating the model uncertainty decomposition.

The primary input is Sentinel-1 Extra-Wide (EW) Ground-Range-Detected Synthetic Aperture Radar (SAR) imagery, using HH and HV backscatter and local incidence angle at 40\,m pixel spacing. Sentinel-1 C-band SAR provides high-resolution, all-weather observations of sea-ice surface and structural properties. Scenes are thermal-noise corrected, radiometrically calibrated, and reprojected to the NSIDC (National Snow and Ice Data Center) Sea-Ice Polar Stereographic North grid (EPSG:3413). Auxiliary inputs include four AMSR2 passive-microwave brightness-temperature channels at 18.7 and 36.5 GHz in horizontal and vertical polarization. Passive-microwave brightness temperatures respond to differences in surface emissivity and physical temperature, providing contrast between open water, sea ice, and snow-covered ice. The polarization and frequency dependence also carries information about ice surface state, snow cover, and melt conditions. Coarser auxiliary variables are aligned to the SAR grid and standardized channel-wise.
Each Sentinel-1 acquisition is paired with temporally valid, spatially overlapping operational ice charts. We rasterize all charts to the common Sentinel-1 grid, allowing pixelwise comparison without registration error. Matching the agency archives yields 404 multi-expert scenes from 2025, whose spatial and seasonal patterns in inter-agency label differences are summarized in Fig.~\ref{fig:atlas}.

\begin{table}[t]
\caption{Inter-expert comparison on shared pixels after label harmonization. Scene counts denote scene--pair combinations rather than distinct scenes. Harm. mism. denotes the fraction of pixels with differing harmonized labels, and $W_1$ measures the magnitude of label differences in tenths.}
\label{tab:pairs}
\centering
\begin{tabular}{lrrr}
\toprule
Pair & Scenes & Harm. mism. & Mean $W_1$ \\
\midrule
DMI--NIC  & 231 & 0.512 & 0.80 \\
CIS--NIC  & 94  & 0.859 & 0.85 \\
NIC--NOAA & 75  & 0.908 & 0.89 \\
CIS--NOAA & 57  & 0.071 & 0.08 \\
CIS--DMI  & 26  & 0.379 & 0.56 \\
\bottomrule
\end{tabular}
\end{table}

\begin{figure}[t]
\centering
\includegraphics[width=\columnwidth]{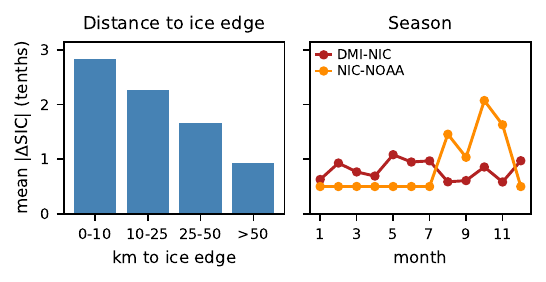}
\caption{Inter-agency SIC differences across 404 scenes.
Left: mean absolute SIC difference as a function of distance to the ice edge.
Right: monthly variation in mean absolute SIC difference for the DMI--NIC and
NIC--NOAA agency pairs.}
\label{fig:atlas}
\end{figure}

\subsection{Experimental Setup}
\label{sec:setup}

All methods use the same U-Net encoder--decoder architecture with four
downsampling stages, channel widths $32/64/128/256$, a 256-channel bridge,
bilinear upsampling, and skip connections. The methods differ only in their
output heads and training objectives. Models are trained on random
$256\times256$ crops with batch size 16 for 30 epochs of 400 iterations using
AdamW with a learning rate of $3\times10^{-4}$, weight decay of $10^{-4}$,
and cosine learning-rate annealing. The architecture, crop sampling,
optimization schedule, and random seeds are fixed across methods, with no
method-specific hyperparameter tuning. Model selection is based on the lowest validation MAE relative to the consensus midpoint. Each experiment is repeated using three random seeds, with the resulting three models also serving as deep-ensemble members for estimating model uncertainty. Experiments are implemented in PyTorch 2.6 with
CUDA 12.4 and run on NVIDIA RTX PRO 6000 GPUs. The scenes are split into 282 training, 41 validation, and 81 test scenes.

\subsection{Baselines}
\label{sec:baselines}

We organize the baselines into four groups according to how they represent
multiple expert labels and interval targets during training.
The first group consists of label-aggregation baselines. These methods use an
11-bin SIC prediction and differ in how the expert labels are combined.
The Hard baseline uses the midpoint of one reference expert label, while
Consensus uses the mean midpoint across the available experts. The Soft
baseline follows label-distribution
learning~\cite{geng2016label,peterson2019human} and averages the harmonized
expert distributions as
$\bar p(j)=\frac{1}{A}\sum_{a\in\mathcal{A}(x)}p_a(j)$.
The second group contains interval-distribution baselines. These methods retain
the expert-reported intervals and represent the prediction with a single
continuous Beta distribution. They differ in how the expert intervals enter
the loss: AvgNLL averages the interval negative log-likelihood across experts,
Cons. uses the consensus range spanning the available expert intervals.
The third group contains deterministic interval-regression baselines.
The limiting loss(LIM)~\cite{cheng2023weakly} gives zero loss when the point
prediction lies inside the reported interval and penalizes deviations outside
it. The Min--max method (MM)~\cite{pukdee2025learning} minimizes the worst-case
squared error over values within the interval. Unlike the distributional
methods, these baselines produce a single point prediction.
Finally, we include a distribution-matching control and two $K=2$ variants.
The $K=1$ model uses the same multi-expert target and Cram\'er objective as the
$K=2$ mixture but has no between-component variation, isolating the effect of
the mixture representation. The base $K=2$ model learns the decomposition from
the interval target alone, while a second $K=2$ variant adds decomposition matching with $\lambda=1$.

\subsection{Metrics}
\label{sec:metrics}

Point accuracy is evaluated using mean absolute error (MAE) against the
consensus midpoint, $\mathrm{MAE}=\frac{1}{N}\sum_{i=1}^{N}
|\hat y_i-\bar y_i|$, where $\hat y_i$ is the predictive mean and $\bar y_i$
is the mean midpoint of the available expert labels. MAE is reported in SIC
tenths.
For probabilistic predictions, predictive entropy is defined as
$H=-\sum_j p(j)\log p(j)$ on the common 11-bin SIC representation.
Distributional accuracy is measured using the
Cram\'er/expected-CRPS score in~\eqref{eq:cramer}, where lower values indicate
closer agreement with the multi-expert target.
Associations between predicted uncertainty and observed quantities are measured
using Spearman rank correlation,
$\rho_s(u,z)=\mathrm{Corr}(\mathrm{rank}(u),\mathrm{rank}(z))$.
We report $\rho(u,W_1)$ to measure association with observed expert
disagreement. For the uncertainty decomposition, we additionally report
$\rho(s_w,t_w)$ and $\rho(s_b,t_b)$ to measure correspondence between the
predicted and observed within- and between-label uncertainty, respectively.
For transfer experiments, $\rho_{\mathrm{tot}}=\rho(u_{\mathrm{tot}},W_1)$
and $\rho_b=\rho(u_b,t_b)$ denote the corresponding total- and
between-component associations.
Finally, selective prediction is evaluated using the excess area under the
error-retention curve (E-AURC)~\cite{geifman2017selective}. Let
$\mathcal{S}_u(c)$ contain the fraction $c$ of predictions with the lowest
uncertainty. The retained error is
$R_u(c)=\frac{1}{|\mathcal{S}_u(c)|}\sum_{i\in\mathcal{S}_u(c)}
|\hat y_i-\bar y_i|$, and
$\mathrm{E\text{-}AURC}=\int_0^1
[R_u(c)-R_{\mathrm{oracle}}(c)]\,dc$, where the oracle retains predictions in
increasing order of their actual error.

\subsection{Main Results}
\label{sec:results}

\begin{table*}[t]
\caption{Point accuracy across supervision regimes and input sets: MAE
(tenths) mean $\pm$ SD over three seeds on test data. }
\label{tab:mae}
\centering
\scriptsize
\setlength{\tabcolsep}{2.6pt}
\begin{tabular}{lcccccccccc}
\toprule
& \multicolumn{3}{c}{Label aggregation}
& \multicolumn{2}{c}{Interval distribution}
& \multicolumn{2}{c}{Interval regression}
& \multicolumn{3}{c}{Distribution matching} \\
\cmidrule(lr){2-4}
\cmidrule(lr){5-6}
\cmidrule(lr){7-8}
\cmidrule(lr){9-11}
Input set
& Hard & Cons. & Soft
& AvgNLL & Cons.
& LIM & MM
& $K=1$ & \textbf{Ours ($K=2$)} & \;\textbf{+ decomp.} \\
\midrule

SAR (3 ch)
& 2.221\pmm{.098}
& 2.129\pmm{.080}
& 1.879\pmm{.065}
& 1.951\pmm{.138}
& 1.820\pmm{.038}
& 2.201\pmm{.154}
& 1.862\pmm{.277}
& 1.484\pmm{.108}
& 1.714\pmm{.126}
& \textbf{1.230\pmm{.083}} \\

SAR+AMSR2 (7 ch)
& 0.809\pmm{.040}
& 0.891\pmm{.162}
& 0.769\pmm{.093}
& 0.862\pmm{.033}
& 0.641\pmm{.039}
& 0.694\pmm{.005}
& 0.732\pmm{.061}
& 0.709\pmm{.012}
& 0.567\pmm{.055}
& \textbf{0.559\pmm{.045}} \\

\bottomrule
\end{tabular}
\end{table*}

Table~\ref{tab:mae} compares how input information and learning strategy affect point accuracy. The rows use SAR (3 channels) and SAR+AMSR2
(7 channels), while the columns compare label aggregation, interval
distribution methods, deterministic interval regression, and distribution
matching with $K=1$ and $K=2$ components. Smaller MAE indicates better
accuracy. 
Retaining richer label information generally improves performance. Within
label aggregation, soft labels outperform hard labels in both input settings,
reducing MAE from $2.221$ to $1.879$ for SAR and from $0.809$ to $0.769$ for
SAR+AMSR2. The interval-based consensus distribution further improves on
aggregation-based supervision ($1.820$ and $0.641$), indicating that
preserving the interval structure of the annotations, rather than collapsing
them to a single target, benefits point accuracy. The deterministic LIM and
MM baselines are weaker than distribution matching when AMSR2 is available
($0.694$ and $0.732$ vs.\ $0.567$ for $K=2$ and $0.559$ with decomposition),
and, unlike the mixture-based models, they do not yield predictive
distributions that support subsequent uncertainty analysis.
The benefit of additional mixture components depends on the input: $K=2$
outperforms $K=1$ with SAR+AMSR2 ($0.567$ vs.\ $0.709$) but underperforms
it with SAR alone ($1.714$ vs.\ $1.484$). The proposed decomposition
resolves this: it recovers and surpasses the $K=1$ result for SAR ($1.230$,
the lowest SAR-only error) and further reduces the SAR+AMSR2 MAE to
$0.559$, the lowest in the table. Decomposition supervision thus stabilizes the mixture when the input alone
cannot disambiguate the components, which explains its larger gain in the
more ambiguous SAR-only setting.
Finally, adding AMSR2 consistently improves all methods, reducing MAE by
approximately 52--69\% relative to the SAR-only models (e.g., $2.221$ to
$0.809$ under hard supervision), confirming that the passive microwave
channels carry substantial complementary information independent of the
learning strategy.

Figure~\ref{fig:mixture} illustrates why a single component is insufficient
when expert annotations are separated. The target consists of two disjoint
expert intervals with an unsupported region between them. Minimizing the
Cram\'er objective forces a single Beta to spread mass across this gap
($0.61$), whereas the two-component mixture preserves both modes, reduces
the gap mass to $0.20$, and lowers the CRPS by $19\times$. The CDF view
(bottom) makes the mechanism explicit: only the mixture can reproduce the
flat plateau of the target CDF over the unsupported region. This failure
mode is exactly the regime addressed by the $K{=}2$ model in
Table~\ref{tab:mae}.
\begin{figure}[t]
\centering
\includegraphics[width=\columnwidth]{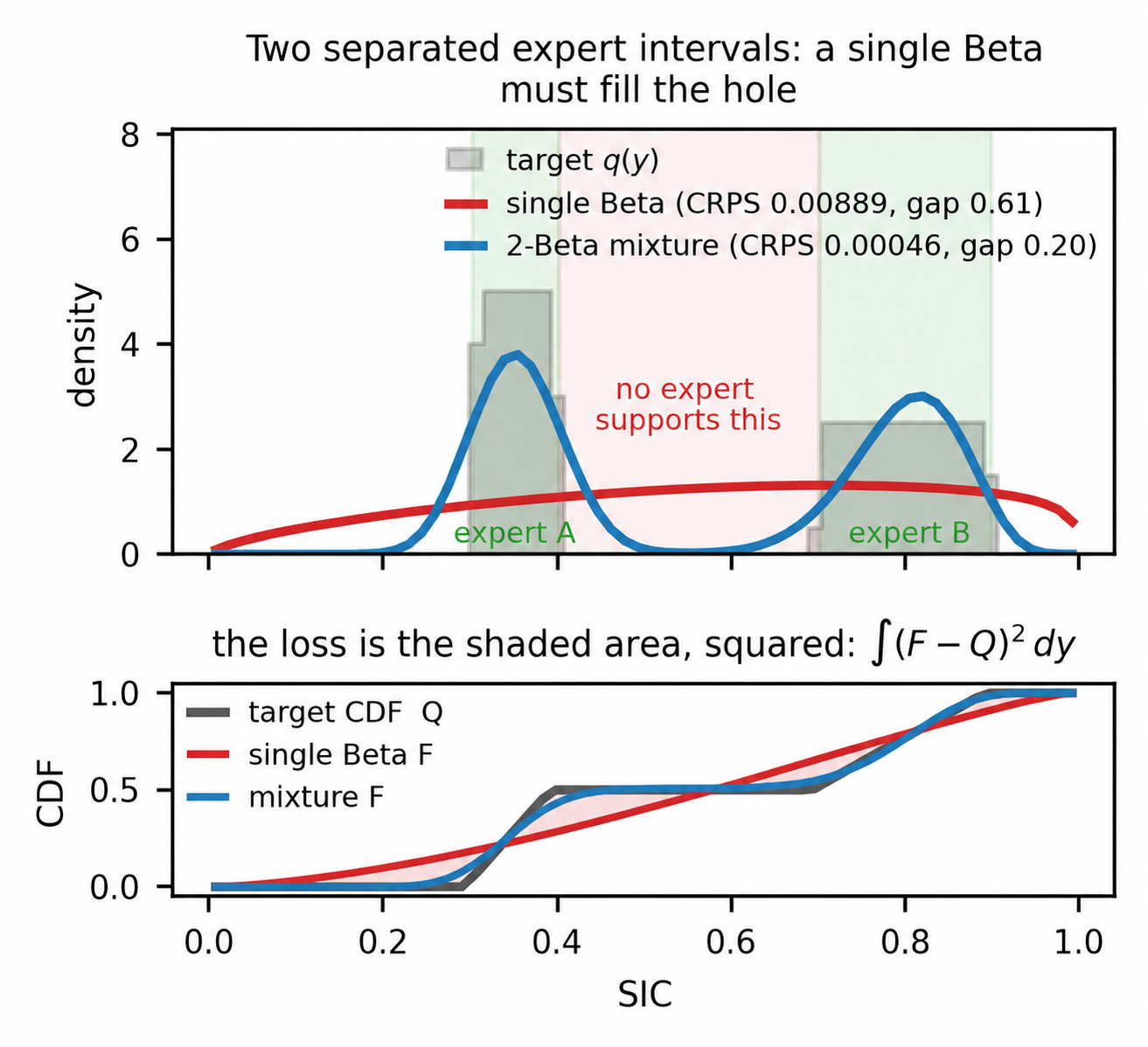}
\caption{Why one component is insufficient for separated labels.
Both models use the same target and Cram\'er objective. A single Beta fills
the unsupported gap between the two agency intervals, whereas the mixture
preserves the two modes and reduces CRPS by 19$\times$.}
\label{fig:mixture}
\end{figure}

\begin{table}[t]
\caption{Predictive uncertainty alignment with $W_1$ and model error under the
7-channel setting. $H$ denotes tenth-bin predictive entropy and $u$
predictive width. $\rho(H,\mathrm{err})$ and E-AURC use $H$ for ranking in all
methods.}
\label{tab:beta}
\centering
\setlength{\tabcolsep}{4.5pt}
\begin{tabular}{lcccc}
\toprule
Method
& $\rho(H,W_1)$
& $\rho(u,W_1)$
& $\rho(H,\mathrm{err})$
& E-AURC \\
\midrule

soft
    & 0.283\pmm{.164}
    & 0.127\pmm{.149}
    & 0.512\pmm{.370}
    & 0.256\pmm{.099} \\
hard
    & 0.015\pmm{.037}
    & 0.021\pmm{.024}
    & 0.747\pmm{.013}
    & 0.095\pmm{.012} \\
consensus
    & 0.075\pmm{.064}
    & 0.009\pmm{.071}
    & 0.744\pmm{.045}
    & 0.099\pmm{.025} \\

AvgNLL
    & 0.073\pmm{.004}
    & 0.039\pmm{.019}
    & \textbf{0.831\pmm{.009}}
    & \textbf{0.082\pmm{.003}} \\
Cons.-interval
    & 0.268\pmm{.073}
    & 0.099\pmm{.077}
    & 0.366\pmm{.204}
    & 0.248\pmm{.067} \\
$K=1$
    & 0.144\pmm{.066}
    & 0.141\pmm{.070}
    & 0.443\pmm{.214}
    & 0.204\pmm{.070} \\
$K=2$
    & 0.364\pmm{.017}
    & 0.200\pmm{.137}
    & 0.135\pmm{.289}
    & 0.336\pmm{.037} \\
$K=2$ + decomp.
    & \textbf{0.397\pmm{.022}}
    & \textbf{0.402\pmm{.026}}
    & 0.177\pmm{.196}
    & 0.309\pmm{.007} \\
\bottomrule
\end{tabular}
\end{table}

Table~\ref{tab:beta} evaluates what predictive uncertainty captures under the SAR+AMSR2 input setting by comparing its correlation with $W_1$, its correlation with model error, and its selective-prediction performance (E-AURC).
Alignment with $W_1$ is strongest for the mixture-based models, and decomposition strengthens it further. The proposed
$K{=}2{+}\mathrm{decomp}$ model achieves the highest correlations with
$W_1$ for both entropy ($\rho(H,W_1)=0.397$) and predictive width
($\rho(u,W_1)=0.402$), doubling the width–$W_1$ correlation of the
plain $K{=}2$ model ($0.200$) and reducing its variance across seeds. In contrast, hard and consensus categorical supervision yield uncertainty that is essentially uninformative about $W_1$ ($\rho(H,W_1)=0.015$ and $0.075$, respectively), which is expected because these targets collapse inter-expert label variation during training.
The pattern reverses for error alignment. The methods with the weakest correlations with $W_1$---AvgNLL, hard, and consensus---show the
strongest correlation between entropy and error
($0.831$, $0.747$, and $0.744$) and the best selective-prediction
performance, with AvgNLL attaining the lowest E-AURC ($0.082$). The distribution-matching models ($K{=}1$, $K{=}2$, $K{=}2$+decomp.), whose uncertainty better reflects variation in the expert labels, correlate only weakly with error ($0.135$--$0.443$) and rank
examples less effectively (E-AURC $\geq 0.204$).
Together, these results indicate that uncertainty aligned with expert-label variation and uncertainty aligned with model error are complementary rather than interchangeable: objectives that preserve inter-expert label variation yield uncertainty that better reflects label ambiguity, whereas objectives that collapse this variation can yield uncertainty that is more effective for ranking prediction errors. The decomposed model is thus preferable when uncertainty is intended to represent ambiguity in the supervision, whereas AvgNLL is preferable when uncertainty is used to identify predictions likely to have large errors.

\begin{table}[t]
\caption{Validation of the three-way uncertainty decomposition on test data
(SAR+AMSR2). Entries are Spearman $\rho$; bold indicates the intended
reference. $u_m$ is computed across seeds.}
\label{tab:decomp_val}
\centering
\setlength{\tabcolsep}{2.5pt}
\begin{tabular}{llcccc}
\toprule
Model & Comp. & $\rho(\cdot,t_w)$ & $\rho(\cdot,t_b)$ & $\rho(\cdot,d_{\rm src})$ & $\rho(\cdot,\mathrm{err})$ \\
\midrule
$K{=}2$
 & $u_w$ & 0.667\pmm{.013} & 0.500\pmm{.008} & 0.366\pmm{.012} & 0.070\pmm{.281} \\
 & $u_b$ & 0.066\pmm{.431} & 0.107\pmm{.291} & 0.149\pmm{.174} & 0.635\pmm{.152} \\
 & $u_m$ & $-$0.206 & $-$0.054 & 0.050 & 0.689 \\
\midrule
$K{=}2$+D
 & $u_w$ & 0.677\pmm{.014} & 0.521\pmm{.015} & 0.392\pmm{.015} & 0.170\pmm{.167} \\
 & $u_b$ & 0.616\pmm{.049} & 0.500\pmm{.018} & 0.396\pmm{.014} & 0.233\pmm{.110} \\
 & $u_m$ & $-$0.266 & $-$0.118 & 0.015 & 0.098 \\
\bottomrule
\end{tabular}
\end{table}
Table~\ref{tab:decomp_val} compares each predictive component with four
references: $t_w$ measures within-label uncertainty, $t_b$ between-expert
variation, $d_{\rm src}$ direct expert disagreement, and $\mathrm{err}$
prediction error. For plain $K{=}2$, $u_w$ aligns well with $t_w$
($0.667\pm0.013$), but $u_b$ aligns weakly with $t_b$ and $d_{\rm src}$
($0.107$ and $0.149$) and instead tracks error ($0.635$), as the Cram\'er
objective constrains only the overall CDF; $u_m$ also tracks error
($0.689$). With decomposition, $u_b$ becomes substantially
better aligned with $t_b$ and $d_{\rm src}$ ($0.500$ and $0.396$) while its
error correlation falls to $0.233$, and $u_m$ loses its error association
($0.098$) as ensemble members converge. Cross-correlations remain, showing
improved source alignment rather than complete disentanglement.

\begin{table}[t]
\caption{Transfer performance under cross-agency and temporal shifts.
 A dash indicates that no between-component
uncertainty channel is available.}
\label{tab:transfer}
\centering
\setlength{\tabcolsep}{2.6pt}
\begin{tabular}{@{}lccc@{}}
\toprule
Method & MAE & $\rho_{\rm tot}$ & $\rho_b$ \\
\midrule
\multicolumn{4}{@{}l}{\textit{Cross-agency transfer: train DMI+NIC, test CIS/NOAA}} \\
\midrule
pair holdout, $K{=}2$
    & 0.727\pmm{.139} & 0.150\pmm{.022} & 0.197\pmm{.025} \\
pair holdout, $K{=}2$ + DS
    & 0.684\pmm{.268} & 0.098\pmm{.029} & 0.112\pmm{.053} \\

\midrule
\multicolumn{4}{@{}l}{\textit{Temporal transfer: train NIC 2018--2024, test 2025}} \\
\midrule
archive hard
    & 1.475\pmm{.109} & 0.321\pmm{.031} & -- \\
archive soft
    & 1.346\pmm{.092} & 0.279\pmm{.027} & -- \\
archive $K{=}2$
    & 0.626\pmm{.053} & 0.409\pmm{.009} & 0.519\pmm{.017} \\
archive $K{=}2$ + decomp.
    & 0.562\pmm{.016} & 0.396\pmm{.008} & 0.528\pmm{.019} \\
\bottomrule
\end{tabular}
\end{table}
Table~\ref{tab:transfer} evaluates point accuracy and uncertainty alignment
under cross-agency and temporal shifts. Here,
$\rho_{\rm tot}=\rho(u_{\rm tot},W_1)$ measures alignment of total predictive
uncertainty with expert disagreement, while
$\rho_b=\rho(u_b,t_b)$ measures alignment of between-component and
between-label uncertainty.
Under temporal transfer (train on NIC charts alone from 2018--2024, test on
the 2025 multi-agency scenes), both $K{=}2$
variants outperform the categorical baselines ($0.626$ and $0.562$ vs.\
$1.475$ and $1.346$), with the decomposed model lowest and most stable
($0.562\pmm{.016}$). The uncertainty channels also transfer: total
uncertainty stays correlated with expert disagreement
($\rho_{\rm tot}\approx0.40$), and the between-component channel aligns
even more strongly with between-label uncertainty ($\rho_b=0.519$ and
$0.528$), a signal unavailable to hard and soft baselines.
Cross-agency transfer is harder: decomposition still lowers MAE ($0.684$
vs.\ $0.727$, within seed variance), but both alignment correlations drop
($\rho_{\rm tot}\leq0.150$, $\rho_b\leq0.197$). Point predictions transfer
to unseen agency pairs, but the uncertainty channels reflect the annotation
conventions of the training agencies.

\begin{table}[t]
\caption{Irreducible MAE and model error by distance to the ice edge (SAR+AMSR2), evaluated against individual agency charts on pixels labeled by $\geq2$ agencies. Ours ($K{=}2$ + decomp.)}
\label{tab:floor}
\centering
\setlength{\tabcolsep}{2.8pt}
\begin{tabular}{lccccc}
\toprule
Stratum & Floor & Hard MAE & Ours MAE & Floor/Hard & Floor/Ours \\
\midrule
pooled
& 0.353 & 0.935\pmm{.057} & \textbf{0.625\pmm{.034}}
& 38\% & \textbf{56\%} \\

0--10 km
& 1.339 & 2.269\pmm{.157} & \textbf{2.191\pmm{.116}}
& 59\% & \textbf{61\%} \\

10--25 km
& 1.041 & 2.008\pmm{.078} & \textbf{1.791\pmm{.048}}
& 52\% & \textbf{58\%} \\

25--50 km
& 0.546 & 1.523\pmm{.093} & \textbf{1.195\pmm{.025}}
& 36\% & \textbf{46\%} \\

50+ km
& 0.461 & 1.453\pmm{.178} & \textbf{1.071\pmm{.018}}
& 32\% & \textbf{43\%} \\

no edge
& 0.180 & 0.598\pmm{.055} & \textbf{0.274\pmm{.034}}
& 30\% & \textbf{66\%} \\
\bottomrule
\end{tabular}
\end{table}

Table~\ref{tab:floor} compares the irreducible MAE estimated from inter-expert label variation (floor) with observed model error across distance-to-ice-edge strata. The
floor is strongly stratified: it is highest immediately at the ice edge
($1.339$ within $0$--$10$~km) and decays monotonically with distance
($0.180$ with no edge present), mirroring the inter-agency label-difference pattern in
Fig.~\ref{fig:atlas} and confirming that irreducible error concentrates
where annotators themselves diverge.
The proposed model attains lower MAE than hard supervision in every
stratum and operates closer to the floor throughout, raising the
Floor/MAE ratio from $38\%$ to $56\%$ pooled. The gains are largest where
the error is most reducible: far from the edge, the decomposed model
recovers much of the headroom left by hard supervision, whereas near the
edge both models already sit close to the floor ($59\%$ and $61\%$ within
$0$--$10$~km), suggesting that the remaining error is largely attributable to label ambiguity rather than model deficiency.
These strata separate two error regimes: away from the edge, errors are
largely reducible and the proposed supervision reduces them; at the edge, inter-expert label variation imposes a larger irreducible error floor, leaving less room for improvement through additional modeling.

\begin{table}[t]
\caption{Synthetic identifiability study. noise replaces the
feature with pure noise. 
}
\label{tab:synth}
\centering
\begin{tabular}{lccccc}
\toprule
&  & \multicolumn{2}{c}{$\rho(s_b,s)$}
& \multicolumn{2}{c}{$s_b^2/(s_w^2+s_b^2)$} \\
\cmidrule(lr){3-4}
\cmidrule(lr){5-6}
$\sigma_s$ & Ceiling &
$K{=}2$ & + decomp. &
$K{=}2$ & + decomp. \\
\midrule
0
& 1.000
& 1.000\pmm{.000}
& 1.000\pmm{.000}
& 0.783
& 0.832 \\

0.20
& 0.591
& 0.589\pmm{.007}
& 0.589\pmm{.006}
& 0.666
& 0.917 \\

\textsc{noise}
& 0.004
& 0.000\pmm{.009}
& 0.011\pmm{.017}
& 0.618
& 0.949 \\
\bottomrule
\end{tabular}
\end{table}
Table~\ref{tab:synth} tests whether between-component uncertainty recovers latent between-expert variation. Both variants remain at the ceiling across noise levels, while the noise control gives near-zero correlation. Decomposition leaves this correlation unchanged but shifts variance into the between-component channel, whose share increases with $\sigma_s$ ($0.832$--$0.949$) instead of decreasing as in plain $K{=}2$ ($0.783$--$0.618$). Thus, the variance share is meaningful only when interpreted alongside $\rho(s_b,s)$.

\subsubsection{Sensitivity Analysis}
\label{sec:sensitivity}

\begin{table}[t]
\caption{Sensitivity to the number of mixture components under the SAR+AMSR2
input setting. For $K\geq2$, $\rho_b=\rho(s_b,t_b)$ and
$\rho_{\rm err}=\rho(s_b,\mathrm{err})$.}
\label{tab:k_sensitivity}
\centering
\setlength{\tabcolsep}{2.8pt}
\begin{tabular}{cccccc}
\toprule
$K$ & MAE & CRPS & $\rho_b$ & $\rho_{\rm err}$ & $K_{\rm eff}$ \\
\midrule
1 & 0.709\pmm{.012} & 0.0320\pmm{.0012}
  & -- & -- & 1.000\pmm{.000} \\
2 & 0.567\pmm{.055} & 0.0243\pmm{.0010}
  & 0.107\pmm{.291} & 0.635\pmm{.152} & 1.089\pmm{.028} \\
3 & 0.560\pmm{.026} & 0.0249\pmm{.0003}
  & 0.186\pmm{.194} & 0.430\pmm{.246} & 1.105\pmm{.008} \\
4 & 0.554\pmm{.029} & 0.0251\pmm{.0004}
  & 0.212\pmm{.212} & 0.456\pmm{.195} & 1.106\pmm{.019} \\
\bottomrule
\end{tabular}
\end{table}

Table~\ref{tab:k_sensitivity} varies mixture size at seven channels. MAE improves from $K{=}1$ to $K{=}2$ ($0.709$ to $0.567$) and then saturates, while CRPS is best at $K{=}2$ ($0.0243$). For all $K\geq2$, $K_{\rm eff}\approx1.1$, indicating little use of additional components. Although $\rho_b$ increases with $K$, its seed variance remains large, while $\rho_{\rm err}$ peaks at $K{=}2$ ($0.635$). We therefore use $K{=}2$ throughout the paper, as it gives the best CRPS with comparable MAE to larger mixtures.

\begin{table}[t]
\caption{Sensitivity to decomposition weights for the $K=2$
mixture under the SAR+AMSR2 input setting. Here,
$\rho_b=\rho(s_b,t_b)$ and
$\rho_{\rm err}=\rho(s_b,\mathrm{err})$.}
\label{tab:lambda_sensitivity}
\centering
\scriptsize
\setlength{\tabcolsep}{2.5pt}
\begin{tabular}{cccccc}
\toprule
$\lambda_w$ & $\lambda_b$ &
$\rho_b$ &
$\rho(s_w,t_w)$ &
$\rho_{\rm err}$ &
MAE \\
\midrule
0   & 0   & 0.107\pmm{.291}  & 0.667\pmm{.013} & 0.635\pmm{.152}  & 0.567\pmm{.055} \\
0   & 0.3 & 0.353\pmm{.222}  & 0.525\pmm{.130} & 0.155\pmm{.180}  & 0.579\pmm{.036} \\
0   & 1   & 0.491\pmm{.020}  & 0.454\pmm{.046} & 0.053\pmm{.194} & 0.576\pmm{.022} \\
0.3 & 0   & 0.093\pmm{.223}  & 0.403\pmm{.250} & 0.365\pmm{.279}  & 0.616\pmm{.026} \\
1   & 0   & $-$0.092\pmm{.009} & 0.678\pmm{.012} & 0.559\pmm{.076} & 0.618\pmm{.014} \\
0.3 & 0.3 & 0.431\pmm{.066}  & 0.559\pmm{.033} & 0.101\pmm{.191}  & 0.555\pmm{.013} \\
1   & 1   & 0.500\pmm{.018}  & 0.677\pmm{.014} & 0.233\pmm{.110}  & 0.559\pmm{.045} \\
3   & 3   & 0.506\pmm{.031} & 0.672\pmm{.023} & $-$0.193\pmm{.027} & 0.663\pmm{.014} \\
\bottomrule
\end{tabular}
\end{table}

Table~\ref{tab:lambda_sensitivity} varies the decomposition weights. Only
$\lambda_b$ improves between-label alignment ($\rho_b$ $0.107\to0.491$ at
$(0,1)$; $\lambda_w$ alone leaves it near zero), but supervising $u_b$ alone
degrades within-label alignment ($0.667\to0.454$); supervising both restores
it, with $(1,1)$ aligning both components ($0.500$, $0.677$) at MAE
comparable to no supervision ($0.559$ vs.\ $0.567$). Two effects bound the
useful range: $\rho_{\rm err}$ falls as $\rho_b$ rises ($0.635\to-0.193$),
and over-weighting hurts accuracy ($0.663$ at $(3,3)$). $(1,1)$ with only a $0.004$ MAE increase over the best setting $(0.3,0.3)$, well within seed variability; we therefore adopt $(1,1)$.

\section{Conclusion}

We introduce a framework for learning from multiple expert interval labels while distinguishing within-label uncertainty, between-label uncertainty arising from variation across experts, and predictive model uncertainty. By harmonizing heterogeneous label conventions and retaining the full interval supervision, the proposed Mixture-of-Betas model reduced MAE by $45\%$ with SAR alone and $31\%$ with SAR+AMSR2 relative to hard supervision, outperforming single-component and deterministic interval-target baselines. Our results show that predictive uncertainty components do not automatically align with their intended sources. Decomposition matching substantially improved alignment between between-component and between-label uncertainty and increased the width--$W_1$ correlation without degrading point accuracy. Uncertainty aligned with between-expert variation and model error remained complementary rather than interchangeable.
Future work will extend the framework to spatially structured uncertainty, semi-supervised decomposition when multi-expert labels are available for only a subset of scenes, and expert-conditioned modeling of agency-specific disagreement. A further direction is to jointly represent uncertainty associated with between-expert variation and model error within a single predictive distribution.


\section*{Acknowledgment}
This research was funded by the National Science Foundation (NSF) under grant number 2531101.



%
\bibliographystyle{IEEEtran}
\bibliography{references}

\end{document}